\documentclass[runningheads]{llncs}
\usepackage{amsmath}
\usepackage{amsfonts, amssymb}
\usepackage{multirow}
\usepackage{xcolor}
\usepackage{dsfont}
\usepackage{hyperref} 

\usepackage[T1]{fontenc}

\usepackage{graphicx}
\begin{document}
\title{On the Risk of Misleading Reports: Diagnosing Textual Biases in Multimodal Clinical AI}
\titlerunning{Diagnosing Textual Bias in Multimodal Clinical AI}

% David Restrepo, Ira Ktena, Maria Vakalopoulou, Stergios Christodoulidis, Enzo Ferrante
\author{David Restrepo\inst{1} \and
Ira Ktena\inst{2} \and
Maria Vakalopoulou\inst{1} \and
Stergios Christodoulidis\inst{1} \and
Enzo Ferrante\inst{3}
}
\authorrunning{D. Restrepo et al.}

\institute{
MICS, CentraleSupélec - Université Paris-Saclay, France \\
\email{\{david.restrepo, maria.vakalopoulou, stergios.christodoulidis\}@centralesupelec.fr}\and
Google DeepMind, London, UK \\
\email{iraktena@google.com}\\ \and
CONICET, Universidad de Buenos Aires, Argentina\\
\email{eferrante@dc.uba.ar}}
\maketitle              % typeset the header of the contribution
\begin{abstract}
Clinical decision-making relies on the integrated analysis of medical images and the associated clinical reports. While Vision-Language Models (VLMs) can offer a unified framework for such tasks, they can exhibit strong biases toward one modality, frequently overlooking critical visual cues in favor of textual information. In this work, we introduce Selective Modality Shifting (SMS), a perturbation-based approach to quantify a model’s reliance on each modality in binary classification tasks. By systematically swapping images or text between samples with opposing labels, we expose modality-specific biases. We assess six open-source VLMs—four generalist models and two fine-tuned for medical data— on two medical imaging datasets with distinct modalities: MIMIC-CXR (chest X-ray) and FairVLMed (scanning laser ophthalmoscopy). By assessing model performance and the calibration of every model in both unperturbed and perturbed settings, we reveal a marked dependency on text input, which persists despite the presence of complementary visual information. We also perform a qualitative attention-based analysis which further confirms that image content is often overshadowed by text details. Our findings highlight the importance of designing and evaluating multimodal medical models that genuinely integrate visual and textual cues, rather than relying on single-modality signals.

\end{abstract}

\section{Introduction}

Clinical decision-making relies on the integrated analysis of medical images and associated textual or metadata information \cite{multimodal_med}. However, Vision-Language Models (VLMs) often exhibit modality bias on favor of language priors over visual inputs, which can lead to unsafe or misleading predictions. Recent work has shown the limitations of large language models in medical domains such as ophthalmology \cite{bias_llms_oph,llms_oph2} and radiology \cite{bias_llms_rad}, highlighting challenges like hallucinations, shortcut reasoning, and embedded biases. Even advanced multimodal decoders frequently overlook image content in favor of text, sometimes generating plausible yet hallucinated outputs \cite{ref1,ref1_2}. These biases can come from the LLM backbone itself, with models confidently predicting outcomes without relying on image features \cite{ref2}. Similar issues have been observed in unimodal and multimodal settings, such as in counterfactual text edits \cite{ref3} and biasing feature interventions \cite{ref4}, both revealing a tendency to ignore visual evidence.

In medical applications, such biases are particularly concerning. A misleading textual report paired with a pathologic image may cause the model to ignore visual signs of disease. Beyond medicine, multimodal transformers trained on general-domain tasks have been shown to develop shortcut learning behaviors—e.g., answering “two” to all “how many” questions—without analyzing the visual scene \cite{ref5,ref6}. This kind of spurious correlation, often caused by data imbalance or domain mismatch, undermines trust in model outputs. Recent studies also highlight that VLMs can produce high-confidence answers in the absence of meaningful visual content \cite{ref2}, suggesting that these models may not be grounded in the image modality even when such grounding is critical.

To address this, we propose a perturbation-based framework—Selective Modality Shifting (SMS)—that systematically swaps image or text components across samples with opposing labels, enabling us to quantify a model’s reliance on each modality. We apply SMS to six VLMs (four general-purpose \cite{llava,llava2,qwen2vl,llama3,janus} and two medically fine-tuned \cite{llavamed,medgemma}) on MIMIC-CXR \cite{mimic-cxr} and FairVLMed \cite{fairclip} datasets. Our evaluation combines standard performance metrics (such as accuracy, precision, recall, and F1), as well as attention-map inspection \cite{attention_exp1,attention_exp2,multimod_exp}, and first-token calibration via Expected Calibration Error (ECE) \cite{ece}. These experiments allows us to capture not only what the models predict but also where they attend and how confident they are. Results show that text bias persists even when visual cues contradict it, with miscalibration emerging as a secondary symptom of this over-reliance on language.

Our contributions can be summarized as: 1) we introduce a novel perturbation-based technique, \emph{Selective Modality Shifting (SMS)}, to diagnose modality bias in VLMs for medical classification; 2) we conduct extensive tests with multimodal medical data, exposing how strong textual signals can overshadow image-based pathologies; 3) we employ attention-based explainability to show how attention patterns shift under SMS, providing qualitative evidence of this text dominance; and 4) we extend bias analysis to model uncertainty by computing the ECE of first-token probabilities in both clean and SMS conditions, revealing that modality bias is accompanied by systematic over-confidence.

\section{Methodology}

\subsection{Problem Statement}

Let \(\mathcal{D}\) be a dataset comprising samples \((I, T, y)\), where \(I \in \mathcal{I}\) is a medical image (e.g., a chest X-ray from MIMIC-CXR \cite{mimic-cxr} or a scanning laser ophthalmoscopy image from FairVLMed \cite{fairclip}), \(T \in \mathcal{T}\) is an associated textual description (such as radiology reports or clinical notes), and \(y \in \{0,1\}\) is the ground-truth binary diagnosis label (e.g., abnormal vs. normal, or glaucoma vs. healthy). We define a function $f_{\text{VLM}}$:

\[
\hat{y} = f_{\text{VLM}}(I, T),
\]

\noindent where $f_{\text{VLM}}$ generates a binary diagnosis label \(\hat{y} \in \{0,1\}\) by prompting a VLM with an instruction prompt describing the diagnostic tasks, followed by a medical image $I$ and a text description $T$. $f_{\text{VLM}}$ maps the generated answer to a binary label by applying regular expressions (RE). The use of RE instead of the direct model answer is required as, sometimes, models do not answer "Yes" or "No" even if they are prompted to do so, but instead they provide more comprehensive answers likes "Yes, the patient presents abnormal findings" (as exemplified in Figure \ref{fig1}) which require some post-processing.

The central challenge addressed in this work is to evaluate the extent to which these models genuinely integrate visual and textual cues rather than relying on textual priors— even when critical visual information is available. To this end, we propose the \emph{Selective Modality Shifting} (SMS) framework. For a given sample \((I, T, y)\), we generate counterfactual examples by selectively perturbing one modality. Specifically, for a sample with \(y=1\), we construct a perturbed sample \((I, T', y)\) where \(T'\) is taken from a sample with \(y=0\), and similarly, a perturbed sample \((I', T, y)\) by replacing \(I\) with an image \(I'\) from a sample with \(y=0\). The difference in the model’s output between the original and perturbed inputs provides a direct measure of the contribution of each modality to the diagnostic decision, offering insights into how effectively these models balance and integrate both image and textual information in clinical settings. 

\subsection{Selective Modality Shifting}
\label{sec:sms}

\begin{figure}[t!]
\centering
\includegraphics[scale=0.5]{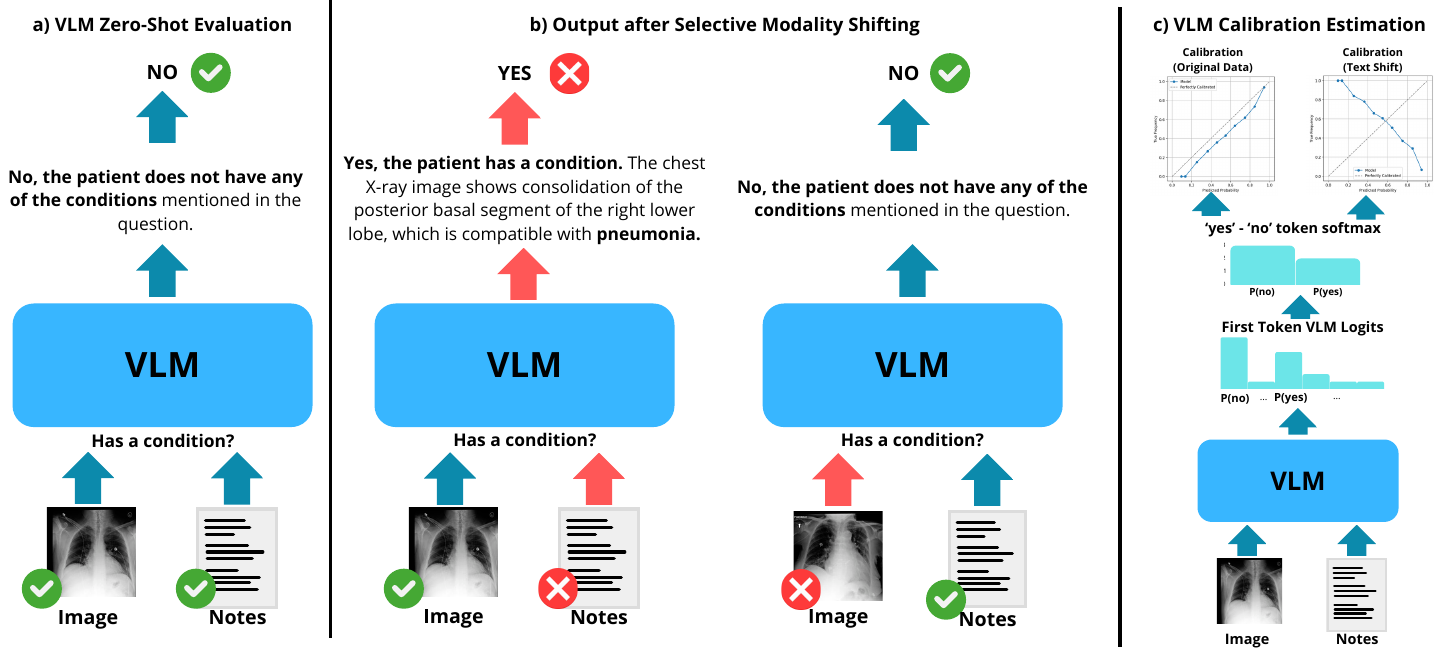}
\caption{Example of Selective Modality Shifting (SMS) and calibration evaluation on chest X-ray: (a) Original inputs yield correct "No condition" output. (b) Text swap induces false "Pneumonia" prediction, unlike image swap—indicating strong textual bias. (c) 
Calibration curves are computed over the probabilities of the tokens "yes" and "no".}
\label{fig1}

\end{figure}

Figure \ref{fig1} provides an illustrative example of SMS. Given a VLM-based function $f_{\text{VLM}}$ that predicts a diagnosis \(\hat{y} \in \{0,1\}\) from an image \(I\) and text \(T\), we systematically replace either the image or the text with a mismatched component from an opposing class. %To isolate the contributions of each modality, we generate two types of perturbed samples from an original triplet \((I, T, y)\), where \(I\) is an image, \(T\) is text, and \(y \in \{0,1\}\) is a binary label. 
By comparing performance metrics (e.g., Accuracy, F1-score) between original and perturbed samples, we quantify the extent of single-modality bias. We hypothesize that extreme sensitivity to either text or image swapping implies over-reliance on that modality.\\

\noindent\textbf{Text Swap.} Let \((I, T, y)\) be a sample with \(y=1\). We select another sample \((I', T', y')\) where \(y' \neq y\) (i.e., \(y'=0\)), and replace only the text to construct a perturbed sample \((I, T', y)\). The model’s prediction for this new sample is
\[
\hat{y}_I^{(T')} = f_{\text{VLM}}(I, T').
\]
We perform this swapping for a number of data samples and measure how large the deviation from the originally predicted labels is by measuring changes in model performance. A large deviation in the predicted labels when only \(T\) is replaced indicates that the model relies heavily on the text modality. Conversely, minimal change suggests that visual features dominate—or that the text is not critical to the model’s reasoning.\\

\noindent\textbf{Image Swap.} Similarly, for the same original sample \((I, T, y)\) with \(y=1\), we select a different sample \((I', T', y')\) where \(y' = 0\) and replace only the image to obtain \((I', T, y)\). The model’s prediction here is
\[
\hat{y}_T^{(I')} = f_{\text{VLM}}(I', T).
\]

If altering the image alone significantly affects the outcome, it implies that the model is integrating visual cues. Otherwise, the model may be underutilizing image information in favor of text.

\subsection{Modality Attention}
\label{sec:attention}
In order to provide a more mechanistic explanation of the model's behavior, we proposed to inspect the attention scores by modality.  Modality attention was calculated for every (image or text) input token. Let \( A_t \in \mathbb{R}^{n} \) be the attention vector for the \( t \)-th generated token over \( n = n_T + n_I \) input tokens, where \( n_T \) and \( n_I \) denote the number of text and image tokens, respectively. We decompose \( A_t \) into modality-specific components: \(A_t = \bigl[A_t^{(T)},\, A_t^{(I)}\bigr]\).

In addition, following \cite{bos_attention}, the attention weight for the beginning-of-sequence (BOS) token is set to zero, i.e., \( A_t(\mathrm{BOS}) = 0 \), to mitigate its confounding effect. This modality attention framework, adapted from \cite{multimod_exp} for Med-LLaVA, provides a precise token-level quantification of the contributions from image and text modalities during generation.

Our goal is to analyze how the attention scores assigned by each output token to the input image and text-tokens change dynamically. In other words, we aim to determine whether different output tokens exhibit distinct attention score patterns for images and text.

\subsection{Calibration Evaluation}
\label{sec:calibration}

As illustrated in Figure~\ref{fig1}, we assess the calibration of first-token probabilities under both original and perturbed conditions. For each image–text pair, we extract the model’s logits for the “yes” and “no” tokens and apply softmax to compute predicted probabilities. While models do not always generate “yes” or “no” as the first token during free-form generation, we find that the probability assigned to these first-token choices remains highly consistent with the model’s full answer. This justifies the use of first-token probabilities as a reliable proxy for binary decisions. We then compute the ECE by binning predictions into 10 uniform intervals and averaging the weighted absolute difference between predicted confidence and empirical accuracy within each bin following the equation:
\[
\text{ECE} = \sum_{m=1}^{M} \frac{|B_m|}{n} \left| \text{acc}(B_m) - \text{conf}(B_m) \right|
\]

\section{Experimental Setup}

\subsection{Datasets}

We analyze two medical imaging datasets, each suited for binary classification ("Yes" vs. "No") based on textual notes and corresponding images. Throughout our experiments, no fine-tuning or additional training is performed. Instead, each model is prompted in a zero-shot manner using images and/or medical reports, relying solely on its existing capabilities. Notably, medical records in both datasets do not always provide a clear diagnosis. In some cases, the text is inconclusive, requiring image analysis to arrive at a final diagnosis.

\paragraph{MIMIC-CXR \cite{mimic-cxr}.} We used 10k random chest radiographs from the test set of MIMIC-CXR repository. Its associated radiology reports are preprocessed to remove uncertain labels and included into a text prompt that asks if the patient’s X-ray is normal or abnormal.

\paragraph{FairVLMed \cite{fairclip}.} This dataset targets glaucoma detection via scanning laser ophthalmoscopy (SLO) images. We used the full test set comprised by 2k images. Metadata and clinical notes are concatenated into a textual prompt that queries for “glaucoma” vs.\ “healthy.” 

\paragraph{Prompting and Tokenization.} In both datasets, we generate short textual templates instructing the model to consider the image and/or text and to provide a diagnosis depending on the dataset. See Figure \ref{fig3} for an example where both text and image are considered (the full set of instruction templates is included in the source code repository available at \url{https://github.com/dsrestrepo/Selective-Modality-Shifting-SMS-}. Tokenization and normalization for the text and image follow the format of each VLM. This setup ensures that all models receive a standardized format, enabling direct comparisons of zero-shot performance.

\subsection{Models \& Metrics}

We evaluate several open-source VLMs from general domain: \emph{LLaVA 1.5}(7B), %\emph{LLaVA 1.6-Mistral}(7B), 
%\emph{Llama3 - LLaVA Next}(8B), 
\emph{Qwen-2 VL}, \emph{Llama 3.2 10B}, %\emph{PaLi-Gamma} (10B), 
\emph{Janus-Pro} (7B), and two from medical domain:
%\emph{Biomed-GPT}, and 
\emph{Med-LLaVA}, and \emph{MedGemma}. The first four are primarily trained on natural-image tasks, whereas the latter two incorporate clinical/biomedical training data. All experiments are conducted on a single A100 GPU with PyTorch, using 16-bit precision (fp16), a temperature of 0, and no sampling for reproducibility.

We measure the classification performance using \emph{Accuracy} (correct predictions over all samples), \emph{Precision} (the proportion of true positive predictions among all predicted positives), \emph{Recall} (the proportion of true positive predictions among all actual positives)  and the \emph{F1-score} (the harmonic mean of precision and recall). These metrics are computed in two settings: (i)~\emph{Unperturbed}, using each dataset as-is, and (ii)~\emph{Perturbed}, where we apply the proposed modality swaps to assess the model’s reliance on text vs.\ image. We also include results for ablations where \emph{only} text or image are fed into the model, directly removing the other modality.

To assess robustness, we adopt the \emph{Negative Flip Rate (NFR)} from~\cite{nfr}, which quantifies the fraction of correct predictions under the base condition \(\hat{y}_i = y_i\) that flip to incorrect under a modality shift \(\hat{y}_i^{(\text{shift})} \ne y_i\). It ranges from 0 (no flips) to 1 (all flipped). Formally:
\[
\text{NFR} = \frac{1}{N} \sum_{i=1}^{N} \mathds{1} \left( \hat{y}_i^{(\text{shift})} \ne y_i,\; \hat{y}_i = y_i \right)
\]

\section{Results and discussion}
\begin{figure}[t!]
\includegraphics[width=\textwidth,scale=1.]{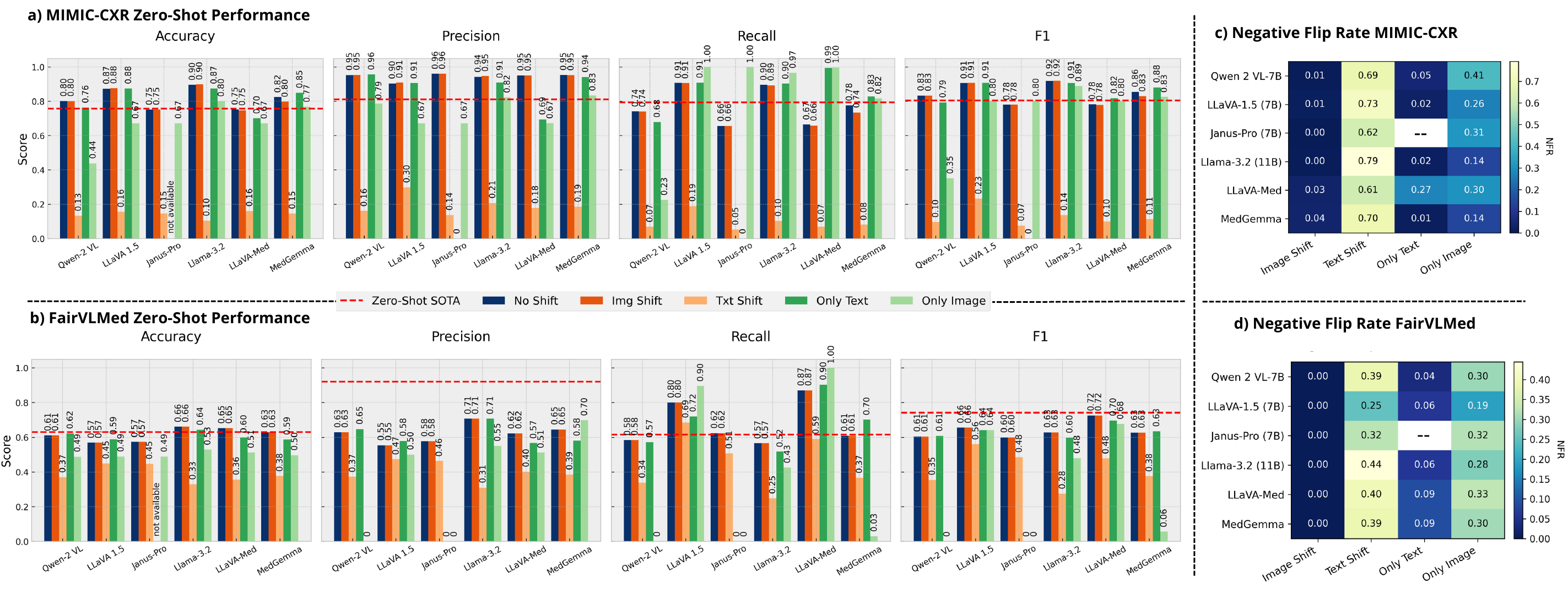}
\caption{Zero-shot classification results on (a)~MIMIC-CXR and (b)~FairVLMed under the Selective Modality Shifting framework. Panels (c) and (d) show the Negative Flip Rate (NFR), measuring how often correct predictions flip to incorrect ones after a perturbation. Janus-Pro does not support text-only input, so the corresponding ablation is omitted. Baseline results from a prior SOTA model are included for reference~\cite{sota}.} \label{fig2}
\end{figure}

Figure \ref{fig2}a,b shows results when applying the SMS framework to MIMIC and FairVLMed. The blue bars (“No Shift”) summarize each model’s performance in the unperturbed scenario. For MIMIC-CXR we observe that the generalist Llama 3 model achieves better accuracy and F1 score, while for FairVLMed, the specialist Med-LLava outperforms overall with a small margin. As a reference, we include results for the same datasets obtained by a SOTA zero-shot model as reported in \cite{sota}. We also compute the NFR (Figure~\ref{fig2}c,d), to quantify how many originally correct predictions are flipped by each perturbation.

\begin{figure}[t!]
\centering
\includegraphics[width=\textwidth]{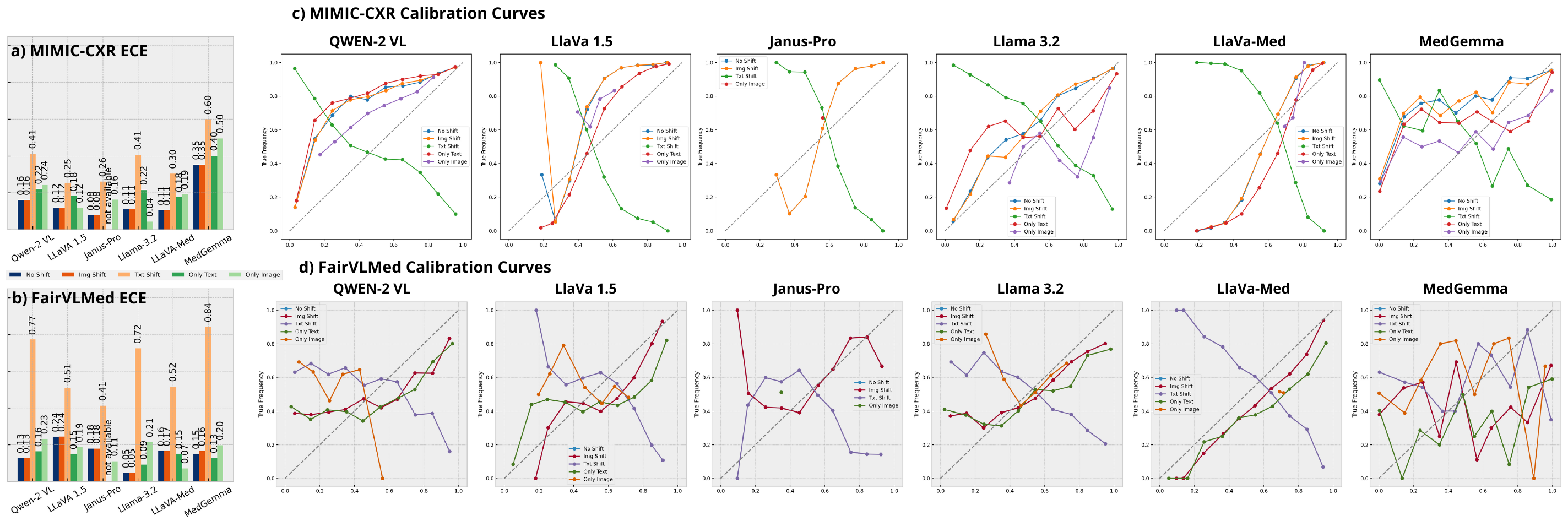}
\caption{(a)~ECE values across shift conditions on MIMIC-CXR and (b)~FairVLMed. (c,d) Calibration curves for all models. A strong misalignment between predicted probabilities and actual frequencies appears under text shifts, reflecting overconfidence.}
\label{fig:calibration}
\end{figure}

Text Shift induces behavior akin to an inverse classifier, where models not only degrade but often flip correct predictions into confident errors, leading to performance far worse than random guessing. For example, in Figure~\ref{fig2}a,c Qwen-2 VL and LLaVA~1.5 lose over 20 points in performance with an NFR of above 0.60 when text alone is perturbed. The \emph{Only Text} and \emph{Only Image} columns further illustrate modality dominance: many models remain functional with text alone yet degrade severely with images alone, confirming the minimal role of visual features in certain predictions. Although domain-specific models like Med-LLaVA display a smaller gap, the overall findings underscore that even clinically tuned VLMs can be susceptible to shortcuts. Additionally, figure~\ref{fig:calibration} shows ECE scores and calibration curves for both datasets. Under text shift, calibration degrades indicating that overconfident errors often coincide with modality-conflicting inputs, supporting our earlier claims on shortcut reliance. This phenomenon is especially pronounced in models like LLaVA-Med and LLaMA 3.2, as illustrated in Figure~\ref{fig:calibration}-c,d, where calibration curves are inverted under the text-shift perturbation.

To assess reliance on a specific modality, we analyze how performance drops when either image (Image Shift) or text (Text Shift) is swapped. We also compute the NFR (see Figure \ref{fig2}c,d), quantifying how many originally correct predictions are flipped by each perturbation. High NFR values—especially under Text Shift—confirm that models heavily depend on textual input. For example, in fig. \ref{fig2}-c Qwen-2 VL has an NFR above 0.75 for text shift, indicating that text mismatches can mislead the prediction.

\begin{figure}[t!]
\centering
\includegraphics[width=\textwidth,scale=0.9]{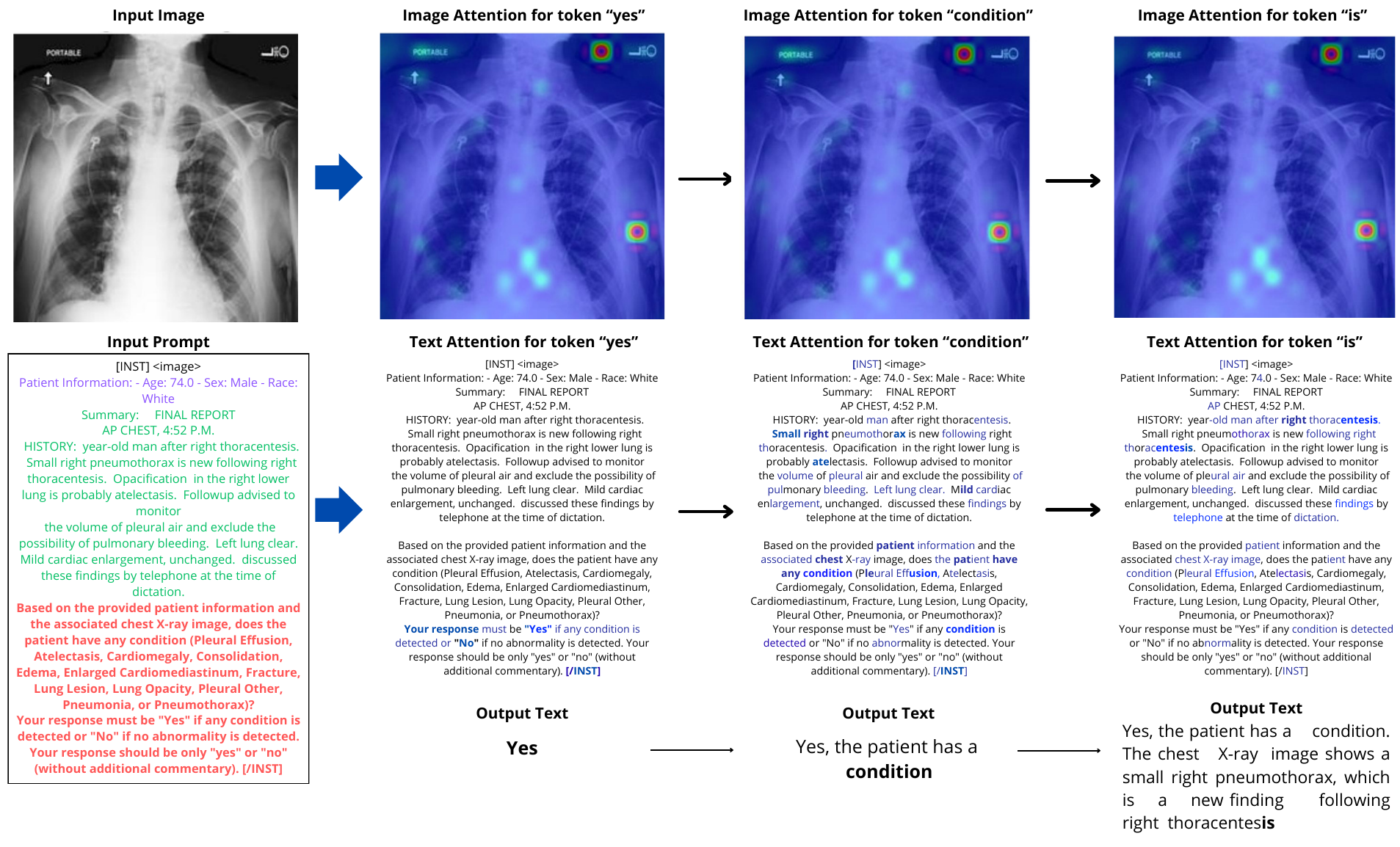}
\caption{Visualization of token-level attention during the generation of an X-ray classification output. The text input prompt is color-coded: \textcolor{violet}{violet} for dataset metadata, \textcolor{green}{green} for clinical notes, and \textcolor{red}{red} for model instructions. Each panel shows how the model distributes attention across input image and text tokens for specific generated tokens ("yes", “condition”, and “is”). The blue intensity of the highlighted text indicates the magnitude of attention directed to each token.} \label{fig3}
\end{figure}

We also performed a qualitative analysis of the attention patterns as discussed in Section \ref{sec:attention}. Due to space limitations, here we present an exemplar case that reflects the tendency observed in our experiments. As shown in Figure \ref{fig3}, the attention patterns for input image tokens remain relatively stable across different output tokens, while for input text tokens vary as new tokens are generated. This observation aligns with our hypothesis that images have minimal influence on the decoding process, whereas text plays a significantly more dominant role.

\section{Conclusions}
In this work, we introduced Selective Modality Shifting, a perturbation-based framework for evaluating the reliance of VLMs on text and image modalities in medical classification tasks. Our findings demonstrate a consistent bias toward textual information, with models frequently disregarding critical visual cues even when presented with complementary image data. This over-reliance on text persists across both general-domain and medically fine-tuned VLMs, raising concerns about the robustness of current multimodal architectures in clinical decision-making.
Through experiments on MIMIC-CXR and FairVLMed, we observed that performance dropped significantly when textual notes were swapped, whereas image perturbations had minimal impact on predictions. Notably, we also found that predictions in these cases were not only less accurate but also poorly calibrated—highlighting overconfidence as a secondary failure mode. These results highlight the need for more rigorous evaluation methodologies to ensure that medical VLMs truly integrate both visual and textual data rather than exploiting shortcut patterns. 

Here we limited our analysis to zero-shot prompting in open models. As future work, we plan to explore if similar conclusions hold for other prompting techniques such as few-shot prompting and more powerful closed models like the GPT or Gemini families.

\begin{credits}
\subsubsection{\ackname} 
This work was partially funded by the European Union’s Horizon Europe programme through the Marie Skłodowska-Curie COFUND grant No. 101127936 (DeMythif.AI).
This work was performed using HPC resources from
the Mesocentre computing center of CentraleSupelec. EF was supported by the Google Award for Inclusion Research and a Googler Initiated Grant. EF and MV are supported by the STIC-AmSud CGFLRVE project.

\subsubsection{\discintname}
The authors have no competing interests to declare that are relevant to the content of this article.

\end{credits}

\end{document}